\documentclass[sigconf]{acmart}

\usepackage{multirow}
\usepackage{makecell}
\usepackage[table]{xcolor}
\AtBeginDocument{%
  }

\setcopyright{acmlicensed}
\copyrightyear{2018}
\acmYear{2018}
\acmDOI{XXXXXXX.XXXXXXX}
\acmConference[Conference acronym 'XX]{Make sure to enter the correct
  conference title from your rights confirmation email}{June 03--05,
  2018}{Woodstock, NY}
\acmISBN{978-1-4503-XXXX-X/2018/06}

\begin{document}

\title{ETPDesigner: Multi-Agent Orchestration for Interactive Multimodal Electronic Theater Program}

\author{Mengtian Li}
\affiliation{%
  \institution{Shanghai University}
  \institution{\mbox{Shanghai Engineering Research Center}}
  \institution{of Motion Picture Special Effects}
  \city{Shanghai}
  \country{China}
}
\email{mtli@shu.edu.cn}

\author{Xinru Guo}
\affiliation{%
  \institution{Shanghai University}
  \city{Shanghai}
  \country{China}
}
\email{simyeoguo@shu.edu.cn}

\author{Xiaoru Lin}
\affiliation{%
  \institution{Shanghai University}
  \city{Shanghai}
  \country{China}
}
\email{23121996@shu.edu.cn}

\author{Xiao Rong}
\affiliation{%
  \institution{Shanghai University}
  \city{Shanghai}
  \country{China}
}
\email{rxqqt777@shu.edu.cn}

\author{Zhifeng Xie}
\affiliation{%
  \institution{Shanghai University}
  \institution{\mbox{Shanghai Engineering Research Center}}
  \institution{of Motion Picture Special Effects}
  \city{Shanghai}
  \country{China}
}
\email{zhifeng\_xie@shu.edu.cn}

\author{Chaofeng Chen}
\authornote{Corresponding author.}
\affiliation{%
  \institution{\mbox{Institute for Math and AI, Wuhan}}
  \institution{School of Artificial Intelligence,}
  \institution{Wuhan University}
  \city{Wuhan}
  \state{Hubei}
  \country{China}
}
\email{chaofengchen@whu.edu.cn}

\renewcommand{\shortauthors}{Li et al.}

\begin{abstract}
  Electronic Theater Programs (ETPs) serve as critical promotional media in the 
  performing arts, comprising a multi-page collection of heterogeneous visual 
  assets such as theatrical posters, performance details, and character 
  portraits. However, existing text-to-image paradigms struggle with such complex 
  design tasks due to their inability to comprehend long-context narratives and 
  maintain visual consistency across multiple distinct pages. To address this, we 
  introduce \textbf{\textit{ETPDesigner}}, a collaborative Multi-Agent framework 
  that directly synthesizes high-quality ETPs from raw dramatic scripts. 
  Emulating a professional design pipeline, our framework orchestrates 
  specialized agents for semantic script analysis, core poster synthesis, 
  functional background generation, and the stratified composition of character 
  assets. Central to ETPDesigner is a global style anchor mechanism that extracts 
  visual priors from the core poster to enforce strict aesthetic uniformity 
  across all generated components. Furthermore, we elevate the ETP from a static 
  publication to an immersive interactive companion. By integrating portrait 
  animation, customized speech synthesis, and persona-grounded Large Language 
  Models (LLMs), our system enables users to engage in real-time, voice-enabled 
  conversations with the generated virtual characters. To rigorously benchmark 
  this task, we construct \textbf{\textit{ETP-Pro}}, a domain-specific benchmark of professional theater posters and high-quality character portraits. 
  Extensive evaluations demonstrate our method's superiority in producing 
  semantically faithful, aesthetically consistent, and highly interactive 
  program sets. 
\end{abstract}


\begin{teaserfigure}
 \includegraphics[width=\textwidth]{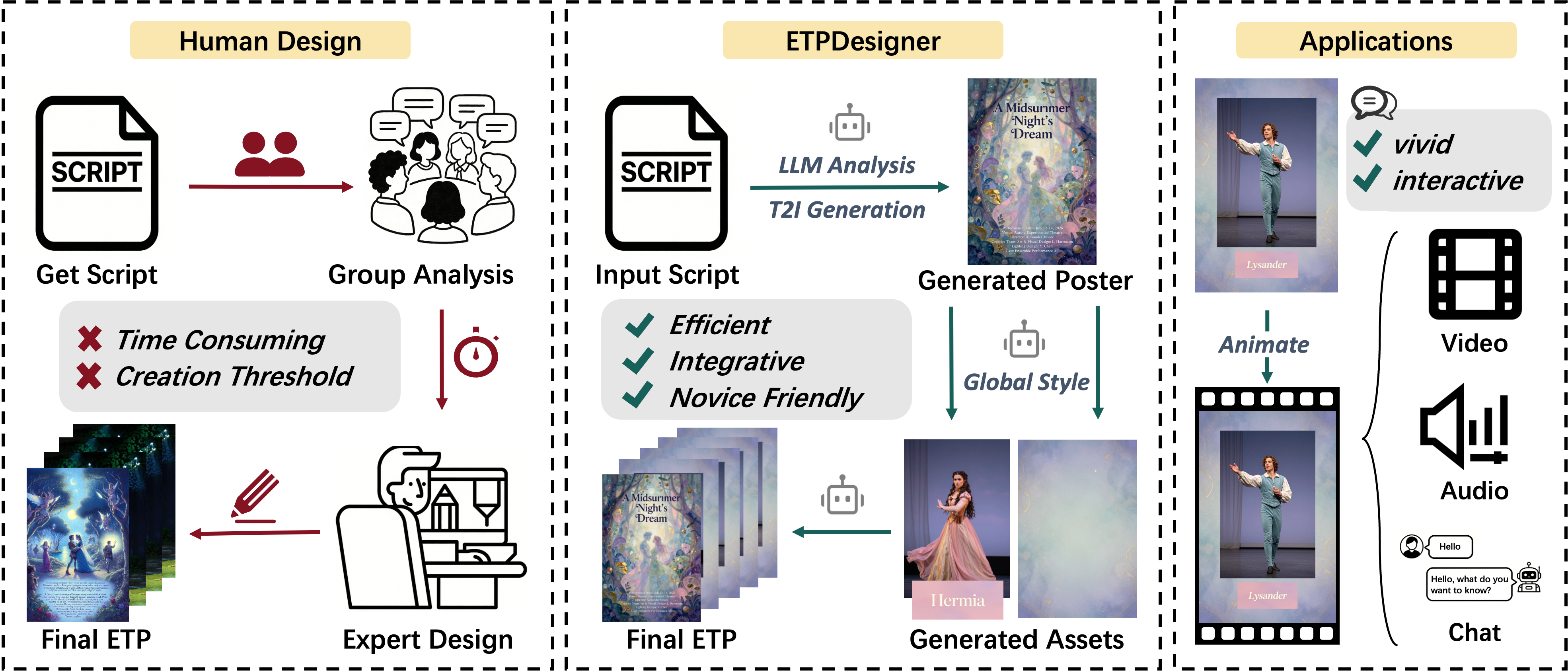}
  \caption{Overcoming the steep threshold of manual design, ETPDesigner automates ETP generation through multi-agent collaboration. It efficiently produces style-unified visual assets that can directly drive real-time interactive character animations.}
  \label{fig:teaser}
\end{teaserfigure}


\maketitle

\section{Introduction}
The Electronic Theater Programs (ETPs) serve as a critical digital medium for the 
promotion and narrative extension of stage arts, including drama, opera, and 
musicals. Unlike cinematic posters that often prioritize realistic celebrity 
close-ups, ETPs demand a higher degree of abstraction and artistic symbolism to 
encapsulate the theatrical theme. Structurally, an ETP acts as a coherent visual 
narrative, sequentially presenting a thematic poster, a synopsis, character 
portraits, and production information. This multi-chapter format distinguishes 
ETP generation from standard slide generation, which focuses purely on layout 
arrangement \cite{li2023gligen, xie2023boxdiff, chen2024training}, and generic 
image editing \cite{brooks2023instructpix2pix, geng2024instructdiffusion, 
sheynin2024emu}, which lacks long-term semantic context. Specifically, the 
synopsis and information sections require background imagery with strategic 
negative space and low-frequency textures to ensure text readability, while 
character portraits necessitate a synthesis of realistic actor features within a 
stylistically consistent atmosphere. Consequently, generating an 
ETP is not merely a task of image synthesis, but a complex orchestration of 
long-context semantic understanding and cross-modal visual consistency.

Recent advancements in Generative Artificial Intelligence (GenAI) have 
revolutionized content creation. However, the professional workflow for 
theatrical design remains labor-intensive, heavily relying on human designers to 
\textbf{interpret textual scripts} and \textbf{manually craft visual assets}. 
While existing automated design 
tools \cite{wei2025postercopilot} offer \textbf{assistance in layout 
composition}, they fundamentally lack the capability to \textbf{comprehend the 
soul of a dramatic script}, including its emotional tone, genre-specific 
aesthetics, and 
character nuances. Therefore, there is an urgent need for an intelligent system 
capable of performing an end-to-end transformation from raw textual scripts to 
fully realized, aesthetically professional visual assets without human intervention.

Despite the capabilities of Large Language Models (LLMs) and Diffusion Models \cite{ho2020denoising, dhariwal2021diffusion}, applying them directly to ETP generation presents three primary challenges. \textbf{(1) General text-to-image models (T2I)} \cite{podell2023sdxl, chen2023pixart, flux2024} \textbf{lack domain-specific aesthetic priors}, struggling to map abstract dramatic themes to appropriate visual design principles without expert guidance. \textbf{(2) Maintaining long-range visual consistency across different functional images is difficult}, where standard generation often results in severe style drift between the main poster and subsequent background or character assets. \textbf{(3) Complex asset-level composition remains a hurdle}, as generating elements that simultaneously satisfy specific character profiles, layout constraints, and background stylistic unity often leads to visual hallucinations in end-to-end generation.

To address these challenges, we propose \textbf{ETPDesigner}, a novel Multi-Agent Collaboration Framework for automated ETP generation. Adopting a training-free paradigm, our system decomposes this complex task into a seamlessly integrated pipeline. Initially, a semantic analysis agent leverages LLMs and expert rules to extract themes and character profiles. Subsequently, to synthesize the main visual poster, we introduce a RAG-enhanced generation loop where a VLM-as-a-Judge optimizes the output via a Tree of Thoughts (ToT) strategy. This poster then serves as a style anchor, from which global stylistic priors are extracted to guide the generation of negative-space backgrounds and decoupled character portraits. These assets are finally assembled into the ETP through a constraint-driven programmatic layout engine. Furthermore, to elevate the user experience beyond static layout synthesis, we introduce an interactive multimodal application. By leveraging portrait animation, customized voice synthesis, and persona-grounded LLMs, the system brings generated characters to life, enabling real-time conversational interactions with users. Recognizing the absence of standardized benchmarks in this emerging domain, we also introduce \textbf{ETP-Pro}, a curated benchmark dataset of dramatic scripts paired with professionally annotated theater assets.

Our contributions can be summarized as follows:
\begin{itemize}
    \item We propose \textbf{ETPDesigner}, the first multi-agent generative 
    framework for automated Electronic Theater Program synthesis, which 
    successfully bridges long-context script understanding with complex visual 
    generation through a novel RAG-enhanced Tree of Thoughts evaluation 
    loop.
    \item We introduce a style-consistent asset-level generation strategy 
    coupled with a programmatic layout engine, and further extend the static 
    program into an immersive multimodal application. This enables real-time, 
    voice-enabled conversational interactions with animated theatrical 
    characters.
    \item We present \textbf{ETP-Pro}, a meticulously curated benchmark dataset 
    comprising dramatic scripts paired with professionally annotated theater 
    posters and character portraits, providing a much-needed benchmark for this 
    emerging domain.
\end{itemize}

\section{Related Work}
\textbf{Graphic Design and Poster Generation.} The automation of graphic design 
has evolved from modular layout engines to holistic, aesthetic-driven synthesis. 
Content-aware pipelines \cite{lin2023autoposter} and hierarchical layout 
protocols optimized via aesthetic feedback \cite{cheng2025graphic, 
wei2025postercopilot} laid the initial groundwork. Recent methods achieve 
seamless compositional control and stylistic harmony by utilizing unified 
multimodal interfaces \cite{dalva2025canvas, zhang2025creatidesign} and 
end-to-end learning \cite{chen2025postercraft, chen2025posta}. Concurrently, 
precise text rendering within complex layouts has been significantly advanced 
through glyph-conditional constraints and character-level embeddings 
\cite{yang2023glyphcontrol, jian2025glyphdraw2, Wang_2025_CVPR}. Despite these 
advancements supported by specialized datasets (e.g., MPDS \cite{DBLP:journals/corr/abs-2410-16840}), existing methods 
remain fundamentally insufficient for generating Electronic Theater Programs. ETPs demand a significantly higher degree of abstract aesthetic reasoning 
and symbolic narrative interpretation, highlighting a critical research gap in 
capturing high-level theatrical abstraction through collaborative generative 
frameworks.

\textbf{Multimodal LLMs for Visual Generation.} The paradigm of visual synthesis has increasingly shifted toward agentic frameworks driven by Multimodal LLMs (MLLMs) \cite{koh2023generating, dong2024dreamllm}. Conceptualized as autonomous agents, these models excel in high-level reasoning, decomposing complex user intentions into executable sub-tasks \cite{wang2024genartist, yang2024idea2img, chen2025t2i}. To enhance generative fidelity, recent works integrate self-correction and reflection mechanisms \cite{wu2024self, suo2025longhorizon}, while leveraging chain-of-thought (CoT) capabilities for global spatial planning and semantic layout synthesis \cite{yang2024mastering, khan2025composeanything}. In long-horizon scenarios, MLLMs enforce narrative consistency \cite{he2025dreamstory, dinkevich2025story2board} and manage full-lifecycle professional design processes \cite{Liu2026posterverse, zhao2025utdesign}. Collectively, the synergy between MLLM reasoning and diffusion-based synthesis provides a robust foundation for complex, multi-stage visual generation tasks requiring both semantic depth and structural precision.

\textbf{Controllable Synthesis and Visual Consistency.} Maintaining identity and 
visual coherence across sequences remains a fundamental challenge in generative 
modeling. Recent methodologies have advanced from basic subject-driven generation 
to complex narrative synthesis. Techniques employing consistent self-attention, 
latent-space priors, and LLM-guided diffusion have proven highly effective in 
preserving multi-subject identities across diverse frames 
\cite{liu2024intelligent, zhou2024storydiffusion, yang2025seed, he2025dreamstory}. For specialized formats like manga and storyboards, structured 
attribute mapping and multi-turn interactive refinement facilitate precise 
control over character poses and panel-to-panel coherence \cite{wu2025diffsensei, 
wang2025autostory, cheng2024autostudio, dinkevich2025story2board}. Additionally, 
self-reflection mechanisms have been introduced to enforce logical consistency in 
long-horizon tasks \cite{suo2025longhorizon}. However, while these approaches 
excel at maintaining object- or character-centric persistence within linear 
narratives, they fall short when applied to Electronic Theater Programs. 
Unlike sequential storytelling, ETPs demand a broader, abstract consistency that 
aligns thematic, atmospheric, and stylistic elements across distinct multimodal 
assets.

\section{ETPDesigner}

Given a raw theatrical script $S$, our objective is to synthesize a comprehensive electronic theater program $P$. We formulate this automated generation task as a training-free mapping function $G: S \rightarrow P$. To ensure both semantic fidelity to the script and high aesthetic quality, $P$ is defined as a multimodal composition:
\begin{equation}
    P = \{I_{post}, I_{info}, I_{char}\}
\end{equation}
where $I_{post}$ serves as the main visual anchor (theatrical poster), $I_{info}$ comprises informative images (e.g., synopsis) synthesized by rendering text onto style-consistent backgrounds ($I_{bg}$), and $I_{char}$ denotes composite character profiles.

Directly bridging the modality gap between a lengthy textual script and an abstract visual publication is a highly ill-posed problem. To address this, we propose a multi-agent collaborative framework that decouples the complex generation process into specialized cognitive and execution roles. Our framework dynamically orchestrates a system of six distinct agents, denoted as $\mathcal{A}$:
\begin{equation}
    \mathcal{A} = \{A_{sem}, A_{art}, A_{gen}, A_{crit}, A_{style}, A_{comp}\}
\end{equation}
Specifically, the \textbf{Semantic Analyst} ($A_{sem}$) parses $S$ to extract structured priors such as themes and character profiles. Based on these priors, the \textbf{Art Director} ($A_{art}$) formulates visual strategies and generative prompts, which are executed by the \textbf{Visual Generator} ($A_{gen}$) using a robust Text-to-Image (T2I) model. To ensure aesthetic and semantic alignment, a Vision-Language Model (VLM) serves as the \textbf{Critic} ($A_{crit}$) to iteratively evaluate and optimize the visual outputs. Furthermore, the \textbf{Style Extractor} ($A_{style}$) distills global stylistic priors from the finalized $I_{post}$ to guide subsequent generation, while the \textbf{Layout Compositor} ($A_{comp}$) handles the deterministic spatial assembly of generated assets and text.

\begin{figure*}[t]
  \begin{center}
    \centerline{\includegraphics[width=\textwidth]{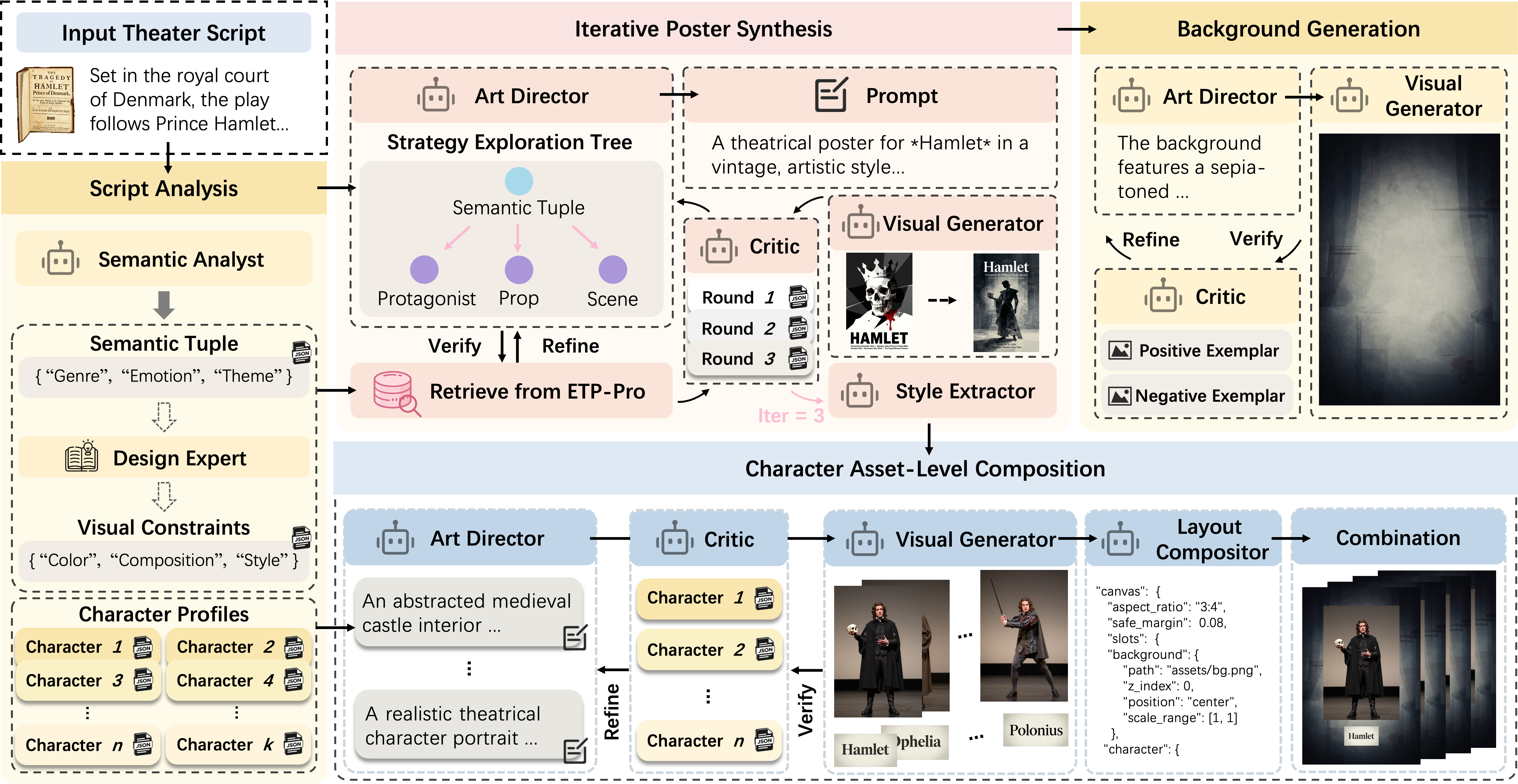}}
    \caption{
      Overview of the ETPDesigner framework. ETPDesigner first extracts 
      semantic profiles and design constraints from the script. 
      It then employs an iterative synthesis loop, 
      where a VLM-as-a-Judge optimizes the visual generation via a Tree of Thoughts strategy. Finally, global stylistic priors are extracted to guide the generation of consistent backgrounds and decoupled character assets, which are assembled into the final ETP through a constraint-driven programmatic layout engine.
    }
    \label{Figure 2}
  \end{center}
\end{figure*}

Driven by $\mathcal{A}$, the mapping function $G$ operates through four collaborative stages: (1) script-to-semantics mapping, (2) self-evaluated and optimized iterative poster synthesis, (3) style-consistent background generation, and (4) asset-level programmatic composition, as illustrated in Figure~\ref{Figure 2}.

\subsection{Script Analysis and Semantic Mapping}

\textbf{Semantic and Character Extraction.} In the initial stage, ${A}_{sem}$ processes the raw input script $\textit{S}$. We 
define a semantic extraction function $\Phi_{\text{LLM}}: \textit{S} \rightarrow 
\textit{H} \times \textit{C}$, which maps the unstructured text into a 
disentangled latent representation. The output comprises two primary components. 
First, a Global Semantic Representation $h \in \textit{H}$, formulated as a tuple 
$h = (g, e, t)$, where $g$ represents the genre classification, $e$ denotes the 
dominant emotional valence, and $t$ is a set of thematic keywords. Second, a 
Character Set $\textit{C} = \{c_1, \dots, c_n\}$. To capture both visual identity 
and theatrical expression, each character profile is formulated as a dual-
component tuple $c_i = (f_{phys}, f_{pers})$. The physical feature set $f_{phys} 
\in \textit{F}_{phys}$ captures morphological details (e.g., costume, age, 
physical build) to guarantee identity consistency across generated assets. 
Meanwhile, the personality feature set $f_{pers} \in \textit{F}_{pers}$ encodes 
abstract psychological traits. Unlike standard extraction paradigms, explicitly 
retaining $f_{pers}$ allows us to condition the generation of unique, character-specific
 stage postures and expressions, thus enhancing the dramatic tension 
of the final portraits ${I}_{char}$.

\textbf{Expert-Guided Semantic Mapping.} To bridge the modality gap and align 
the generative process with theatrical design standards, ${A}_{sem}$ queries a 
heuristic expert system. We formalize this as a deterministic mapping function 
$\Psi_{\text{expert}}: \textit{H} \rightarrow \textit{V}$, which translates the 
high-level semantic representation $h$ into a concrete set of visual design 
constraints $\textit{V} = \{v_{color}, v_{comp}, v_{style}\}$. This mapping is 
strictly grounded in theatrical design theory:
\begin{equation}
    v_{color} = \psi_{col}(e); 
    v_{comp} = \psi_{lay}(g); 
    v_{style} = \psi_{tex}(t)
\end{equation}
where $v_{color}$, $v_{comp}$, and $v_{style}$ dictate the emotion-driven color 
palette, the genre-based spatial composition strategy, and the theme-specific 
rendering style, respectively. For instance, the mapping $\psi_{col}$ might 
translate a negative emotional valence ($e$) into a low-saturation, cool-toned 
color scheme. By conditioning all downstream generation tasks on $\textit{V}$, we 
guarantee that the synthesized visual elements inherently satisfy domain-specific 
aesthetic rules without requiring additional model fine-tuning.

\subsection{Poster Synthesis with Self-Refinement}


\textbf{Strategy Formulation and RAG Retrieval.} Based on the structured 
semantic priors (theme, genre, emotional tone) extracted by ${A}_{sem}$, 
${A}_{art}$ functions as the root of our ToT structure. It devises a visual 
strategy by selecting the most suitable main visual element from a candidate 
space $\mathcal{E} = \{\text{protagonist}, \text{symbolic prop}, \text{key 
scene}\}$. Combined with the design rules from the expert system, ${A}_{art}$ 
synthesizes a highly descriptive prompt $p^{(0)}$, driving ${A}_{gen}$ to 
produce the initial poster $I^{(0)}$. Concurrently, to ground the evaluation in 
professional aesthetics, we employ a Retrieval-Augmented Generation (RAG) module. 
Given the script, it queries the ETP-Pro dataset to retrieve the three most 
semantically similar professional theatrical posters, denoted as the reference 
set ${R}_{ref}$. 

\textbf{Dual-Level Evaluation and ToT Routing.} Both $I^{(0)}$ and 
${R}_{ref}$ are fed into ${A}_{crit}$ for a dual-level assessment. 
(1) \textbf{\textit{Strategy Pruning.}} ${A}_{crit}$ first 
    cross-references $I^{(0)}$ with ${R}_{ref}$ to determine if the selected 
    visual element strategy is appropriate. If the strategy is deemed flawed, 
    the current ToT branch is pruned. ${A}_{crit}$ returns strategic feedback 
    to ${A}_{art}$, prompting it to switch to an alternative element in 
    $\mathcal{E}$ and regenerate a new branch. (2) \textbf{\textit{Visual Optimization.}} If the strategy is validated, 
    the ToT progresses to the refinement node. ${A}_{crit}$ evaluates the 
    image across three dimensions comprising eight sub-metrics (scored 1-10): 
    \textit{Theatrical Aesthetics} (abstraction, tension, refinement), 
    \textit{Communication Efficacy} (integration, clarity), and \textit{Thematic 
    Fidelity} (resonance, aptness, contextuality).

\textbf{Iterative Refinement Logic.} The dimension score is the average of its 
sub-metrics. We introduce a threshold-based optimization logic: if any 
sub-metric scores below 6.5, ${A}_{crit}$ mandates executing the optimization 
suggestions for that specific dimension. Conversely, if all sub-metrics pass the 
6.5 threshold, the system is forced to optimize the dimension with the lowest 
average score. Based on the targeted 
feedback $f$, ${A}_{art}$ updates the prompt to $p^{(k)}$, and ${A}_{gen}$ 
synthesizes the refined image $I^{(k)}$. This inner loop iterates up to a 
maximum of $K=3$ times.

\textbf{Final Selection and Style Extraction.} Upon reaching the iteration limit 
$K$, the system evaluates the entire trajectory of generated images. It selects 
the poster with the highest cumulative evaluation score among the initial and all 
refined candidates. This decisive selection process is formalized as:
\begin{equation}
    {I}_{post}^* = \arg\max_{I \in \{I^{(0)}, \dots, I^{(K)}\}} \text{Score}(I)
\end{equation}
Finally, ${A}_{style}$ analyzes ${I}_{post}^*$ to distill 
the global visual style vector, denoted as $\textit{s}_{global} = (s_{art}, 
s_{color})$, where $s_{art}$ represents the overall artistic style and 
$s_{color}$ captures the dominant color palette. This extracted vector 
$\textit{s}_{global}$ acts as an immutable stylistic prior, ensuring that the 
subsequent background and character profile generation stages strictly inherit 
the poster's aesthetic tone.

\subsection{Style-Consistent Background Generation}
\label{sec3.3}

\textbf{Constraint-Driven Synthesis.} To ensure global aesthetic coherence, ${A}_{art}$ initializes the generation process conditioned on the 
global style vector $\textit{s}_{global} = (s_{art}, s_{color})$ extracted in 
the previous stage. ${A}_{art}$ synthesizes a specialized prompt $p_{bg}^{(0)}$ 
that fuses these stylistic priors with functional layout constraints (e.g., 
``highly abstract,'' ``expansive negative space''). ${A}_{gen}$ then executes this prompt to produce the initial background 
candidate $I_{bg}^{(0)}$.

\textbf{Contrastive Evaluation via VLM.} To rigorously assess the functional 
viability of the generated background, we introduce a contrastive evaluation 
mechanism driven by ${A}_{crit}$. We construct a bipartite 
reference set containing a positive exemplar $E^+$ (an abstract image with 
dominant negative space, ideal for typography) and a negative exemplar $E^-$ 
(a visually cluttered image with salient foreground objects that obscure text 
readability). ${A}_{crit}$ evaluates the current candidate $I_{bg}^{(t)}$ by 
performing a cross-attention comparison against both $E^+$ and $E^-$.

\textbf{Feedback and Iterative Refinement.} The evaluation yields a usability 
judgment and actionable textual feedback $f_{bg}$. If $I_{bg}^{(t)}$ exhibits 
excessive visual saliency or fails to provide sufficient text-rendering space 
(thus aligning closer to the traits of $E^-$), the validation fails. 
${A}_{crit}$ explicitly dictates the direction for improvement (e.g., 
``reduce central object saliency,'' ``increase edge blurring''). 
${A}_{art}$ incorporates this feedback to refine the prompt into
$p_{bg}^{(t+1)}$, instructing ${A}_{gen}$ to synthesize a corrected 
image. This contrastive feedback loop iterates until ${A}_{crit}$ 
determines that the image structurally aligns with the functional traits of 
$E^+$, successfully yielding the final style-consistent background 
${I}_{bg}$.

\subsection{Asset-Level Generation and Composition}

Directly generating character layouts with precise typographic elements via end-to-end T2I models frequently introduces spatial distortions and text rendering artifacts. To circumvent these limitations, we propose an asset-level synthesis and programmatic composition strategy, collaboratively executed by the agent suite $\{A_{art}, A_{gen}, A_{crit}, A_{comp}\}$.

\textbf{Decoupled Prompting and Asset Synthesis.} For each character $c_i = \{f_{phys}, f_{pers}\} \in C$, where $f_{phys}$ and $f_{pers}$ denote physical and personal attributes respectively, $A_{art}$ formulates a decoupled bipartite prompt. It establishes a global environmental prior $p_{env}$ based on the script's setting, and derives a specialized character prompt $p_{char}^i$. The combined prompt is concatenated to synthesize the initial character portrait via the generator $A_{gen}$:
\begin{equation}
    p^i = p_{env} \oplus p_{char};
    I_{char}^{(i, 0)} = A_{gen}(p^i)
\end{equation}

\textbf{Profile Verification and Optimization.} To guarantee absolute semantic fidelity to the narrative, $A_{crit}$ evaluates $I_{char}^{(i, 0)}$ against the source profile $c_i$. Functioning as a meticulous proofreader, it detects missing traits or hallucinatory artifacts (e.g., mismatched props or attire). Any detected discrepancy triggers an iterative refinement loop until the visual asset faithfully reflects the persona, yielding the optimized portrait $I_{char}^{i*}$. Concurrently, a standardized typographic nameplate, $I_{name}^i$, is generated, rendering the character's name onto a solid minimalist background.

\textbf{Deterministic Spatial Assembly.} 
Following the preparation of all modular visual assets, the programmatic agent $A_{comp}$ executes the final deterministic assembly. Guided by a master layout template $T_{layout}$, which strictly specifies spatial coordinates, scaling factors, and z-index ordering, $A_{comp}$ integrates the optimized portraits, nameplates, and the style-consistent background $I_{bg}$ (generated in Section \ref{sec3.3}). This procedural integration directly produces the final composite character pages $I_{char}$:
\begin{equation}
    I_{char} = A_{comp}(\{I_{char}^{i*}\}_{i=1}^n, \{I_{name}^i\}_{i=1}^n, I_{bg} \mid T_{layout})
\end{equation}

Unlike non-deterministic generative models, this rule-based compositing guarantees absolute spatial precision and aesthetic uniformity, culminating in the professional digital publication $P$.

\begin{table*}[t] 
  \caption{Quantitative evaluation using VLM-as-a-Judge. We compare our method 
  and its ablation variants against state-of-the-art baselines across 
  multi-dimensional aesthetic and functional metrics. Results are reported as Mean $\pm$ Std across the ETP-Pro.}
  \label{tab1}
  \begin{center}
    \begin{small}
        \renewcommand{\arraystretch}{1.2}
        \newcommand{\sd}[1]{{\scriptsize \color{gray} $\pm$ #1}}
        
        \resizebox{\textwidth}{!}{
        \begin{tabular}{lcccccccccc} 
          \toprule
          \multirow{2}{*}{\textbf{Method}} & \multicolumn{3}{c}{\textbf{Theatrical Aesthetics}} & \multicolumn{2}{c}{\textbf{Communication Efficacy}} & \multicolumn{3}{c}{\textbf{Thematic Fidelity}} & \multirow{2}{*}{\textbf{Expressiveness}} & \multirow{2}{*}{\textbf{Whitespace}}\\ 
          
          \cmidrule(lr){2-4} \cmidrule(lr){5-6} \cmidrule(lr){7-9}
          
           & \textbf{Abstraction} & \textbf{Tension}  & \textbf{Refinement}  & \textbf{Integration}  &  \textbf{Clarity}  & \textbf{Resonance}  & \textbf{Aptness}  & \textbf{Contextuality} & & \\ 
          \midrule
          Nano Banana Pro      & 4.58 \sd{0.85} & 6.25 \sd{0.72} & 6.91 \sd{0.68} & 6.79 \sd{0.74} & 5.33 \sd{0.91} & 5.28 \sd{0.88} & 4.62 \sd{1.05} & 5.07 \sd{0.82} & 3.95 \sd{1.12} & 1.56 \sd{0.65}\\
          Seedream 4.5         & 4.61 \sd{0.82} & 5.98 \sd{0.79} & 7.83 \sd{0.58} & 6.81 \sd{0.72} & 5.87 \sd{0.85} & 5.64 \sd{0.84} & 4.83 \sd{0.98} & 5.58 \sd{0.80} & 3.82 \sd{1.08} & 1.73 \sd{0.68} \\
          AutoGen              & 3.92 \sd{1.15} & 6.13 \sd{0.76} & 8.06 \sd{0.55} & 7.35 \sd{0.62} & 5.64 \sd{0.88} & 5.46 \sd{0.86} & 5.87 \sd{0.79} & 5.61 \sd{0.81} & 2.19 \sd{0.95} & 2.84 \sd{0.88}\\
          \rowcolor{blue!10} \textbf{Ours} & \textbf{7.93} \sd{0.38} & \textbf{6.99} \sd{0.45} & \textbf{8.27} \sd{0.35} & \textbf{7.53} \sd{0.41} & \textbf{8.04} \sd{0.36} & \textbf{7.11} \sd{0.44} & \textbf{7.96} \sd{0.35} & \textbf{7.87} \sd{0.39} & \textbf{8.43} \sd{0.55} & \textbf{8.19} \sd{0.37}\\
          \midrule
          Ours w/o Expert Sys  & 5.34 \sd{0.88} & 6.35 \sd{0.75} & 8.13 \sd{0.48} & 7.16 \sd{0.61} & 5.69 \sd{0.82} & 5.73 \sd{0.85} & 5.69 \sd{0.81} & 5.85 \sd{0.78} & 8.43 \sd{0.55} & 6.28 \sd{0.72}\\
          Ours w/o Critic ${A}_{crit}$ & 7.17 \sd{0.65} & 6.56 \sd{0.68} & 7.97 \sd{0.52} & 7.49 \sd{0.58} & 7.62 \sd{0.55} & 6.58 \sd{0.72} & 7.39 \sd{0.61} & 6.64 \sd{0.70} & 8.03 \sd{0.61} & 7.66 \sd{0.56}\\
          Ours w/o Style Extr  & 7.93 \sd{0.38} & 6.99 \sd{0.45} & 8.27 \sd{0.35} & 7.53 \sd{0.41} & 8.04 \sd{0.36} & 7.11 \sd{0.44} & 7.96 \sd{0.35} & 7.87 \sd{0.39} & 8.43 \sd{0.55} & 7.89 \sd{0.48}\\
          Ours w/o Asset Comp  & 7.93 \sd{0.38} & 6.99 \sd{0.45} & 8.27 \sd{0.35} & 7.53 \sd{0.41} & 8.04 \sd{0.36} & 7.11 \sd{0.44} & 7.96 \sd{0.35} & 7.87 \sd{0.39} & 8.43 \sd{0.55} & 8.19 \sd{0.37}\\
          \bottomrule
        \end{tabular}
        } 
    \end{small}
  \end{center}
\end{table*}

\section{ETP-Pro}

\begin{figure}[h]
  \centering
  \includegraphics[width=\linewidth]{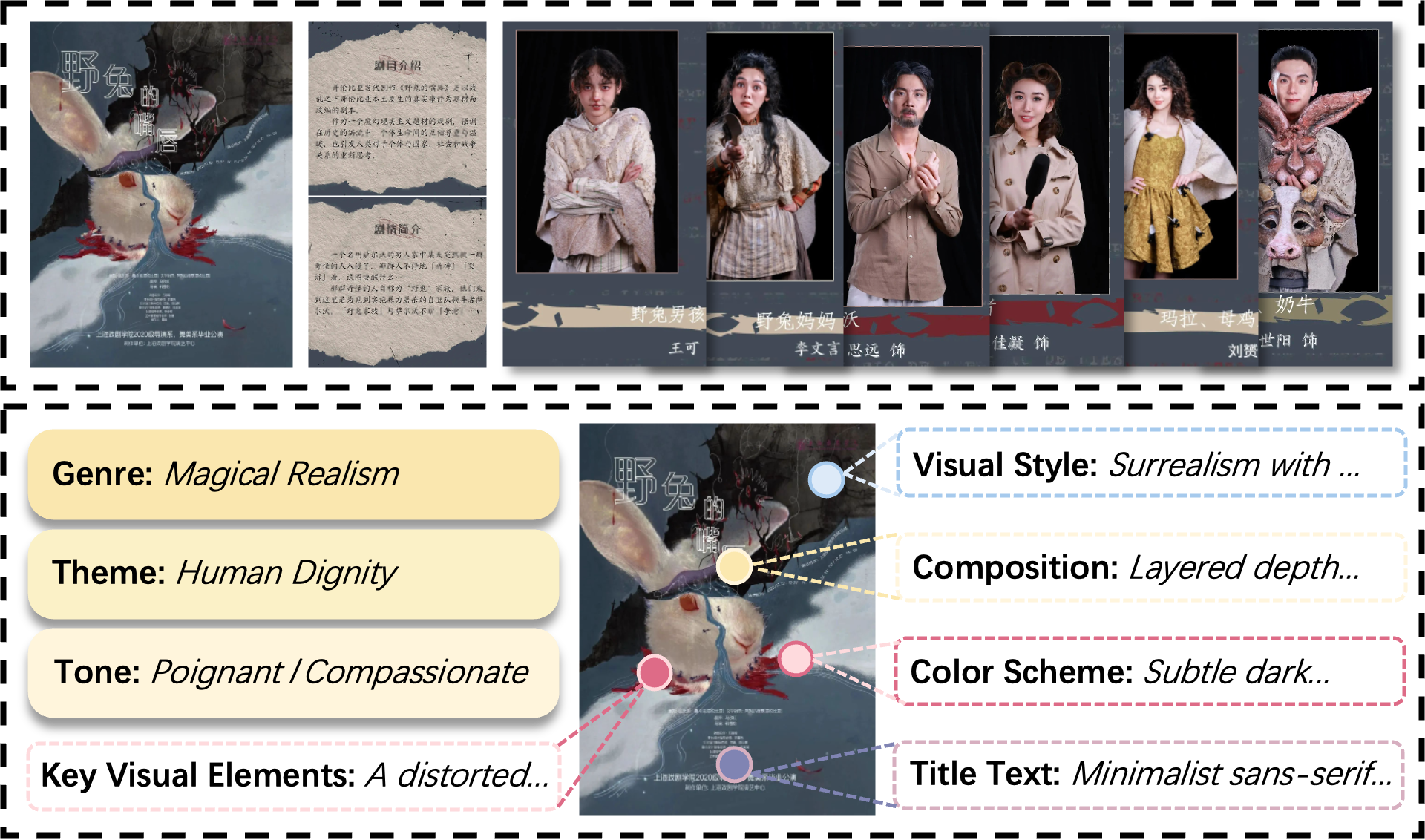}
  \caption{ETP-Pro Benchmark. We curated a collection of existing complete Electronic Theater Programs and conducted detailed annotations on 
  the visual assets.}
  \label{fig3}
\end{figure}

To rigorously evaluate our framework's capabilities in multimodal narrative 
synthesis, we introduce \textbf{\textit{ETP-Pro}}, a domain-specific benchmark specifically curated for Electronic Theater Program generation. ETP-Pro comprises a high-quality collection of 100 unique electronic program. The data composition and annotation methodology of the benchmark are illustrated in Figure~\ref{fig3}.

\textbf{Dataset Statistics and Diversity.} To ensure robust evaluation, the benchmark encompasses a broad stylistic and cultural spectrum. Structurally, the corpus consists of 37 musicals and 63 spoken dramas. Thematically, the entries span 73 dramas, 19 comedies, and 8 tragedies. Geographically, ETP-Pro bridges diverse cultural contexts by including 61 original Chinese works and 39 international productions or adaptations, thereby providing a comprehensive perspective on global theatrical aesthetics.

\textbf{Visual Asset Taxonomy.} Each entry in ETP-Pro strictly corresponds to a source script and contains three distinct asset categories: (1) \textit{Key Visual Posters}, which serve as the primary aesthetic anchors exhibiting high thematic fidelity; (2) \textit{Functional Information Sheets}, which prioritize layout readability through the strategic utilization of negative space; and (3) \textit{Character Portraits}, which reflect script-driven personality traits and specific costume designs.

\textbf{Annotation.} To facilitate fine-grained evaluation, the Key Visual Posters are annotated across three core dimensions: chromatic composition, spatial layout structure, and overall aesthetic style. To ensure both rigorous accuracy and domain-specific professionalism, these annotations are produced via a VLM-assisted generation approach coupled with meticulous expert refinement.

\begin{figure*}[t]
  \begin{center}
    \centerline{\includegraphics[width=\textwidth]{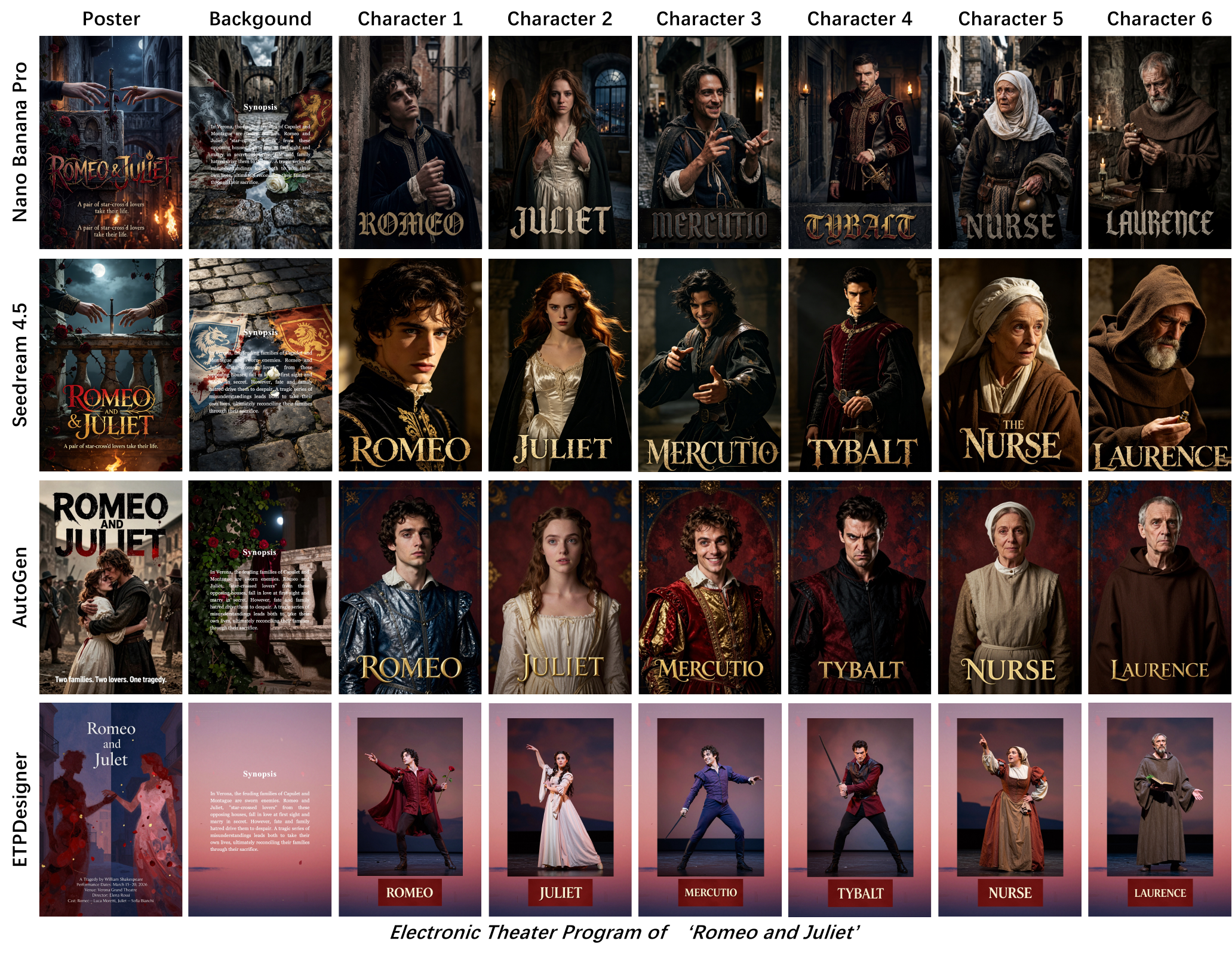}}
    \vspace{-14pt}
    \caption{
      Qualitative comparison between ETPDesigner, Nano Banana Pro, Seedream 4.5, 
      and PosterCraft. The visual results demonstrate that ETPDesigner generates 
      theatrically abstract core posters, functional backgrounds with ample 
      whitespace, and stylistically consistent character portraits, whereas 
      baselines suffer from literal interpretation, layout clutter, and style drift.
    }
    \label{fig4}
  \end{center}
\end{figure*}

\section{Experiments}
\subsection{Experiment Setup}

\textbf{Evaluation Metrics.}  To comprehensively evaluate the generated ETPs, 
we employ a hybrid metric suite comprising VLM-based subjective assessments and objective computational metrics.

\textbf{\textit{VLM-as-a-Judge Metrics.}} Traditional pixel-level metrics struggle to 
capture the abstract nature and functional requirements of theatrical design. 
Therefore, we utilize a state-of-the-art VLM as an impartial judge to evaluate 
the generated assets on a 1-10 scale. The evaluation is systematically 
categorized into asset-specific dimensions: 
For the thematic posters ${I}_{post}$, the evaluation spans three sub-metrics: 
(1) \textit{Theatrical Aesthetics}, which assesses abstraction,
tension, and refinement; 
(2) \textit{Communication Efficacy}, which evaluates the integration of visual 
elements and semantic clarity; 
(3) \textit{Thematic Fidelity}, which measures the resonance and contextuality 
of the image relative to the source script $\textit{S}$. 
For the character portraits ${I}_{char}$, we evaluate:
(4) \textit{Performative Expressiveness}, which assesses whether the characters' poses, gestures, and facial dynamics accurately reflect the exaggerated and dramatic essence inherent to stage performances. 
For the background images ${I}_{bg}$, we measure:
(5) \textit{Functional Negative Space}, which evaluates the strategic reservation of empty areas and low-frequency textures to seamlessly accommodate typographic overlays without causing visual clutter.

\textbf{\textit{Objective Computational Metrics.}} To strictly evaluate the structural and stylistic integrity of the entire publication ${P}$, we utilize the following established metrics:
(1) \textit{Text Accuracy.} This metric evaluates text rendering precision by calculating the Optical Character Recognition (OCR) word recognition rate on the generated textual elements.
(2) \textit{Stylistic Consistency:} We utilize CLIP Image Similarity (CLIP-I) and 
DINOv2 Similarity. Specifically, CLIP-I is computed across all generated visual 
assets within a single ETP set (including the poster ${I}_{post}^*$, background 
${I}_{bg}$, and character portraits ${I}_{char}$) to measure the global 
stylistic coherence. Meanwhile, DINOv2 similarity is calculated pairwise among 
the character portraits to assess fine-grained structural and horizontal artistic 
stability.

\textbf{Baselines.} Given the emerging nature of end-to-end ETP generation, there are currently no existing methods capable of serving as a direct baseline for this complex, multi-asset task. To establish a rigorous and comprehensive evaluation, we carefully design baselines from two distinct perspectives: state-of-the-art T2I models and general multi-agent frameworks. \textbf{(1) Advanced T2I Models (Seedream 4.5} \cite{seedream2025seedream40} \textbf{\& Nano Banana Pro} \cite{deepmind2025nanobanana}\textbf{):} To overcome the context limits of standard T2I models on long scripts, we adopt a two-stage proxy approach. Specifically, an LLM first extracts and recaptions the script into detailed visual prompts, which are then fed into Seedream 4.5 and Nano Banana Pro. This comparison highlights that conventional T2I pipelines, even with LLM preprocessing, struggle to maintain cross-asset consistency without a dedicated framework.
\textbf{(2) General Multi-Agent Framework (AutoGen} \cite{wu2023autogen}\textbf{):} We configure AutoGen with generic agents to simulate our pipeline, demonstrating the superiority of our specialized cognitive roles (e.g., RAG-enhanced Critic). For strict fairness, both AutoGen and ETPDesigner utilize \textbf{Z-Image} \cite{team2025zimage} as the backbone generation model, ensuring that performance differences stem solely from the architectural orchestration rather than raw model capabilities.

\subsection{Qualitative Analysis}

As illustrated in Figure~\ref{fig4}, we visually compare ETPDesigner against
state-of-the-art baselines. While the baselines produce visually appealing 
cinematic posters, they fail to capture the unique abstract aesthetics and 
functional layout requirements of ETPs. In contrast, ETPDesigner successfully synthesizes posters with distinct theatrical abstraction and 
functional backgrounds featuring 
strategic negative space. Furthermore, our method generates highly consistent, 
photorealistic character portraits that seamlessly incorporate narrative-aligned 
props and expressive stage actions. Additional experimental results are detailed 
in the Appendix.

\subsection{Quantitative Evaluation}

To rigorously validate our framework, we compare it against the selected baselines using both VLM-as-a-Judge and objective computational metrics. The results are summarized as follow:

\textbf{VLM-as-a-Judge Analysis.} Qwen3-VL-7B~\cite{Qwen3-VL} evaluations (Table~\ref{tab1}) demonstrate that ETPDesigner consistently secures the highest scores across Theatrical Aesthetics, Communication Efficacy, and Thematic Fidelity. While baselines often produce overly literal or chaotic cinematic outputs, our framework successfully synthesizes stage-aware assets with expressive portraits and strategic functional whitespace. This superiority stems from our multi-agent cognitive planning and iterative optimization, grounding the generative process in professional theatrical aesthetics. We conducted parallel evaluations across all results using GPT-4o \cite{openai2024gpt4ocard} and Gemini-1.5-Pro \cite{geminiteam2024gemini15}, with the detailed outcomes provided in the Appendix.

\textbf{Objective Metrics Analysis.} As reported in Table~\ref{tab2}, ETPDesigner 
achieves the highest CLIP-I and DINOv2 scores, validating the efficacy of our global style anchoring and programmatic assembly in enforcing strict stylistic and structural consistency across heterogeneous assets. Regarding text accuracy, while AutoGen holds a marginal lead, this is merely an artifact of its simplistic generative constraints, as it typically renders only brief titles and lacks essential informational depth. Our method robustly processes highly complex and dense typography, including extensive cast lists and detailed performance schedules, without compromising the underlying visual quality. This demonstrates our framework's superior capability for practical, professional-grade layout generation.

\subsection{Human Evaluation}

\begin{figure}[h]
  \centering
  \includegraphics[width=\linewidth]{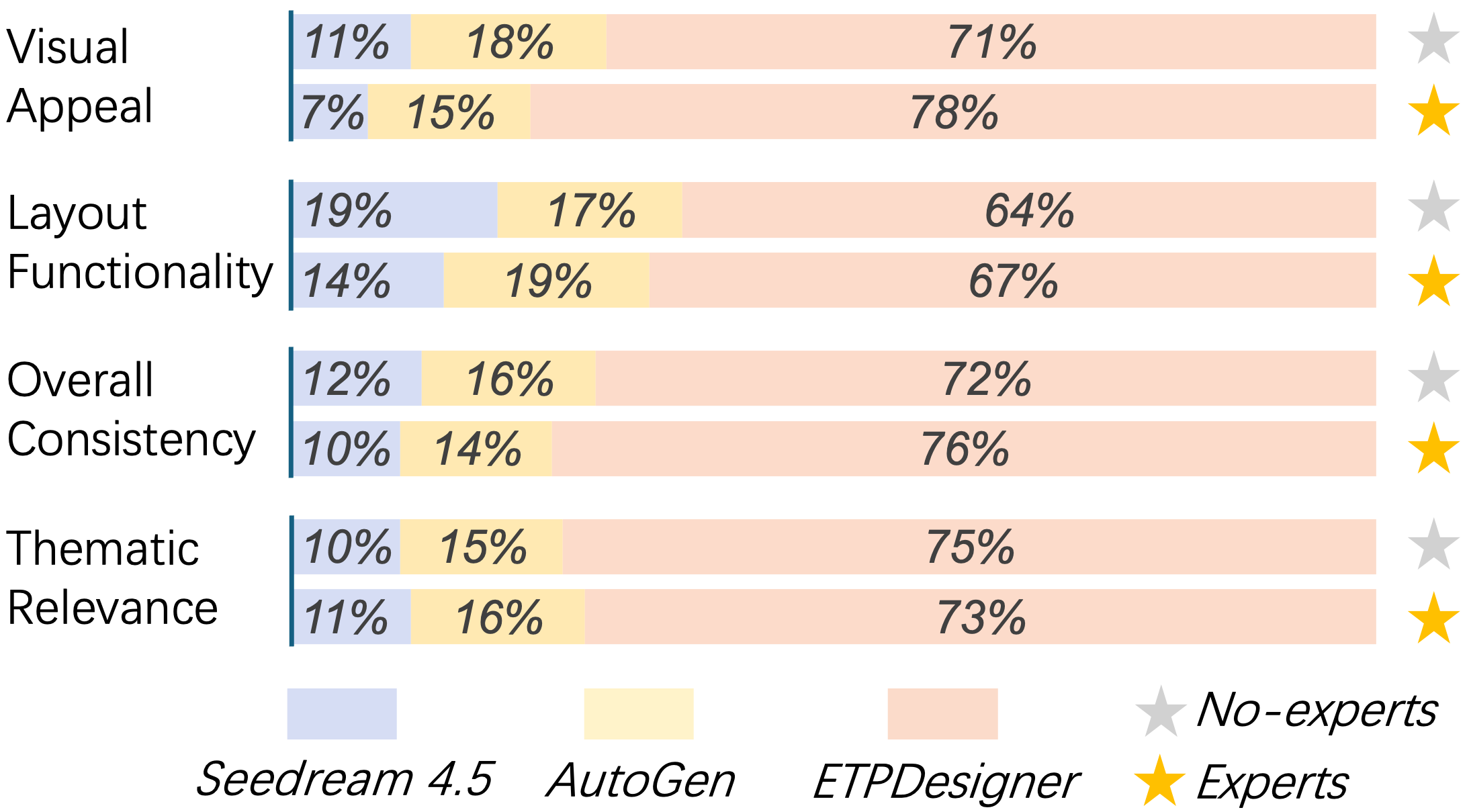}
  \vspace{-15pt}
  \caption{Comparison of Three Methods Across Four Dimensions: 
  (1) Visual Appeal, (2) Layout Functionality, 
  (3) Overall Consistency, (4) Thematic Relevance.}
  \label{fig5}
\end{figure}

To rigorously assess the practical utility of our framework, we conducted a 
comprehensive human evaluation involving 49 participants (12 design experts and 
37 general users). As illustrated in Figure~\ref{fig5}, ETPDesigner 
significantly outperforms the Seedream and AutoGen baselines across four key 
metrics: visual appeal, layout functionality, overall consistency, and thematic 
relevance. These results validate the practical efficacy of our method in 
synthesizing highly aesthetic, structurally robust, and thematically cohesive 
electronic theater programs, securing strong preference from both domain 
professionals and general audiences.

\subsection{Ablation Study}

To validate the necessity and effectiveness of the core components within ETPDesigner, we evaluated several ablation variants, with results detailed in Table~\ref{tab1} and Table~\ref{tab2}.

\textbf{\textit{Effect of the Expert System.}} Removing the cognitive planning agents (Semantic Analyst and Art Director) leads to a substantial decline in Theatrical Aesthetics and Thematic Fidelity (Table~\ref{tab1}). This indicates that without expert-guided prompt formulation, the model fails to capture the abstract dramatic essence, defaulting instead to generic and literal visual interpretations.

\textbf{\textit{Effect of the Critic Agent (${A}_{crit}$).}} The exclusion of the iterative evaluation loop results in noticeable performance drops across both VLM scores and objective metrics. Specifically, the DINOv2 score drops from 0.684 to 0.635, underscoring the critical role of ${A}_{crit}$ in meticulously rectifying semantic hallucinations and structural misalignments. Further results and 
qualitative examples regarding the impact of ${A}_{crit}$ are detailed in the Appendix.

\textbf{\textit{Effect of the Style Extractor (${A}_{style}$).}} Omitting the global style anchoring mechanism causes a severe degradation in cross-asset consistency. As shown in Table~\ref{tab2}, CLIP-I and DINOv2 scores drop sharply to 0.758 and 0.612, respectively. This confirms that explicitly distilling stylistic priors from the core poster is indispensable for harmonizing heterogeneous program pages.

\textbf{\textit{Effect of Asset Composition.}} When replacing our programmatic assembly strategy with direct end-to-end layout generation, the DINOv2 score plummets to 0.528. This massive decline, alongside reduced typography control, validates our decoupled generation-and-composition approach for maintaining precise spatial alignment and professional structural integrity.

\begin{table}[h]
  \caption{Objective Metrics Evaluation. Quantitative comparison of our proposed 
  method, its ablation variants, and baseline models across Text Accuracy, 
  CLIP-I, and DINOv2. Results are reported as Mean $\pm$ Std across the ETP-Pro test set.}
  \label{tab2}
  \begin{center}
    \begin{small}
        \renewcommand{\arraystretch}{1.2}
        \newcommand{\sd}[1]{{\scriptsize \color{gray} $\pm$ #1}}
        
        \begin{tabular*}{\linewidth}{l@{\extracolsep{\fill}}ccc}
          \toprule
          \textbf{Method} & \makecell[c]{\textbf{Text} \\ \textbf{Accuracy $\uparrow$}} & \textbf{CLIP-I $\uparrow$} & \textbf{DINOv2 $\uparrow$} \\
          \midrule
          Nano Banana Pro               & 0.734 \sd{0.12} &  0.625 \sd{0.09} &   0.416 \sd{0.11} \\
          Seedream 4.5                  & 0.867 \sd{0.08} &   0.684 \sd{0.08} &  0.462 \sd{0.10} \\
          AutoGen                       & \cellcolor{blue!10}\textbf{0.897} \sd{0.03} &  0.742 \sd{0.07} &   0.512 \sd{0.09} \\
          \textbf{Ours}                 & 0.831 \sd{0.03} & \cellcolor{blue!10}\textbf{0.857} \sd{0.02} &  \cellcolor{blue!10}\textbf{0.684} \sd{0.03} \\
          \midrule
          Ours w/o Expert System        & 0.831 \sd{0.03} &   0.841 \sd{0.03} &   0.663 \sd{0.05} \\
          Ours w/o Critic ${A}_{crit}$  & 0.802 \sd{0.03} &   0.826 \sd{0.05} &   0.635 \sd{0.06} \\
          Ours w/o Style Extractor ${A}_{style}$ & 0.831 \sd{0.03} &   0.758 \sd{0.08} &   0.612 \sd{0.07} \\
          Ours w/o Asset Composition    & 0.825 \sd{0.03} &   0.814 \sd{0.05} &   0.528 \sd{0.09} \\
          \bottomrule
        \end{tabular*}
    \end{small}
  \end{center}
\end{table}

\section{Applications}

To elevate the user experience beyond static layout synthesis, we introduce an 
interactive multimodal application built upon the generated Electronic Theater 
Program. Specifically, we leverage portrait animation techniques to bring the 
static character assets to life, coupled with customized speech synthesis to 
establish distinct auditory personas for each role. Furthermore, by integrating 
a persona-grounded LLM, the system facilitates real-time, voice-enabled 
conversational interactions. Consequently, users can dynamically engage with 
the virtual roles to inquire about character backgrounds and plot nuances, 
transforming the traditional program into an immersive theatrical companion.

\begin{figure}[h]
  \centering
  \includegraphics[width=\linewidth]{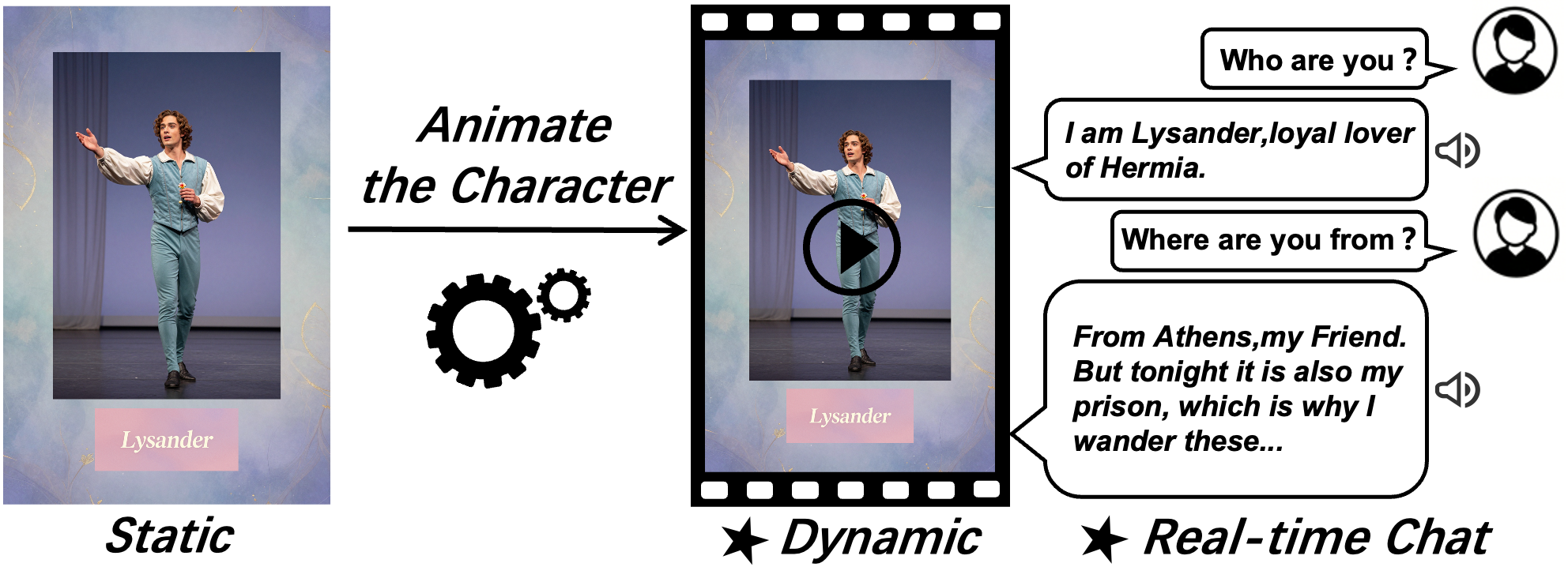}
  \caption{Applications:Transforming static images into character animations and real-time persona-grounded dialogues. }
\end{figure}

\section{Conclusion}

In this work, we introduced ETPDesigner, a multi-agent framework that transforms 
raw dramatic scripts into cohesive Electronic Theater Programs. By decoupling the 
generation pipeline into semantic script analysis, global style anchoring, and 
functional asset synthesis, our approach ensures rigorous theatrical abstraction, 
structural validity, and cross-component stylistic consistency. Crucially, we 
elevate the static generation into an interactive multimodal companion by 
integrating portrait animation, customized speech synthesis, and persona-grounded 
LLMs, enabling real-time, voice-driven engagement with the generated characters. 
In addition to these methodological contributions, we contribute ETP-Pro, a domain-specific benchmark curated from 
professional archives, which serves as a pioneering resource to catalyze future 
research in automated theatrical publicity and interactive design.

\begin{acks}
This work is supported by the National Natural Science Foundation of China (Grant No. 62402306), the Natural Science Foundation of Shanghai (Grant No. 24ZR1422400, Grant No. 25ZR1401130), the Open Research Project of the State Key Laboratory of Industrial Control Technology, China (Grant No. ICT2024B72).
\end{acks}

\bibliographystyle{ACM-Reference-Format}
\bibliography{sample-base}


\end{document}